\documentclass{article}

\usepackage[T1]{fontenc}
\usepackage{amssymb,amsmath}
\usepackage{mathtools}
\usepackage{bm}
\usepackage{esvect}
\usepackage{txfonts}
\usepackage{microtype}
\usepackage{listings}
\usepackage{graphicx}
\usepackage{subfigure} 
\usepackage{natbib}

\usepackage{algorithm}
\usepackage{algorithmic}

\usepackage[nohyperref]{mlp2017}
\usepackage{url}
\usepackage{pgfplots}

\mlptitlerunning{ZSL for Artistic Material Recognition}
\usepackage{verbatim}

\begin{document} 

\twocolumn[
\mlptitle{Multi-Class Zero-Shot Learning for Artistic Material Recognition}

\centerline{\textbf{Alexander W Olson, Andreea Cucu, Tom Bock}}
\centerline{alex.olson@utoronto.ca,  andreea.cucu@gmail.com, tom.henry.bock@gmail.com}
\centerline{All authors contributed equally}

\vskip 7mm
]

\begin{abstract} 
Zero-Shot Learning (ZSL) is an extreme form of transfer learning, where no labelled examples of the data to be classified are provided during the training stage. Instead, ZSL uses additional information learned about the domain, and relies upon transfer learning algorithms to infer knowledge about the missing instances. ZSL approaches are an attractive solution for sparse datasets. Here we outline a model to identify the materials with which a work of art was created, by learning the relationship between English descriptions of the subject of a piece and its composite materials. After experimenting with a range of hyper-parameters, we produce a model which is capable of correctly identifying the materials used on pieces from an entirely distinct museum dataset. This model returned a classification accuracy of 48.42\% on 5,000 artworks taken from the Tate collection, which is distinct from the Rijksmuseum network used to create and train our model.
\end{abstract} 

\section{Introduction}
\label{sec:intro}

\begin{figure}[tb]
\centering
\includegraphics[width=0.4\textwidth]{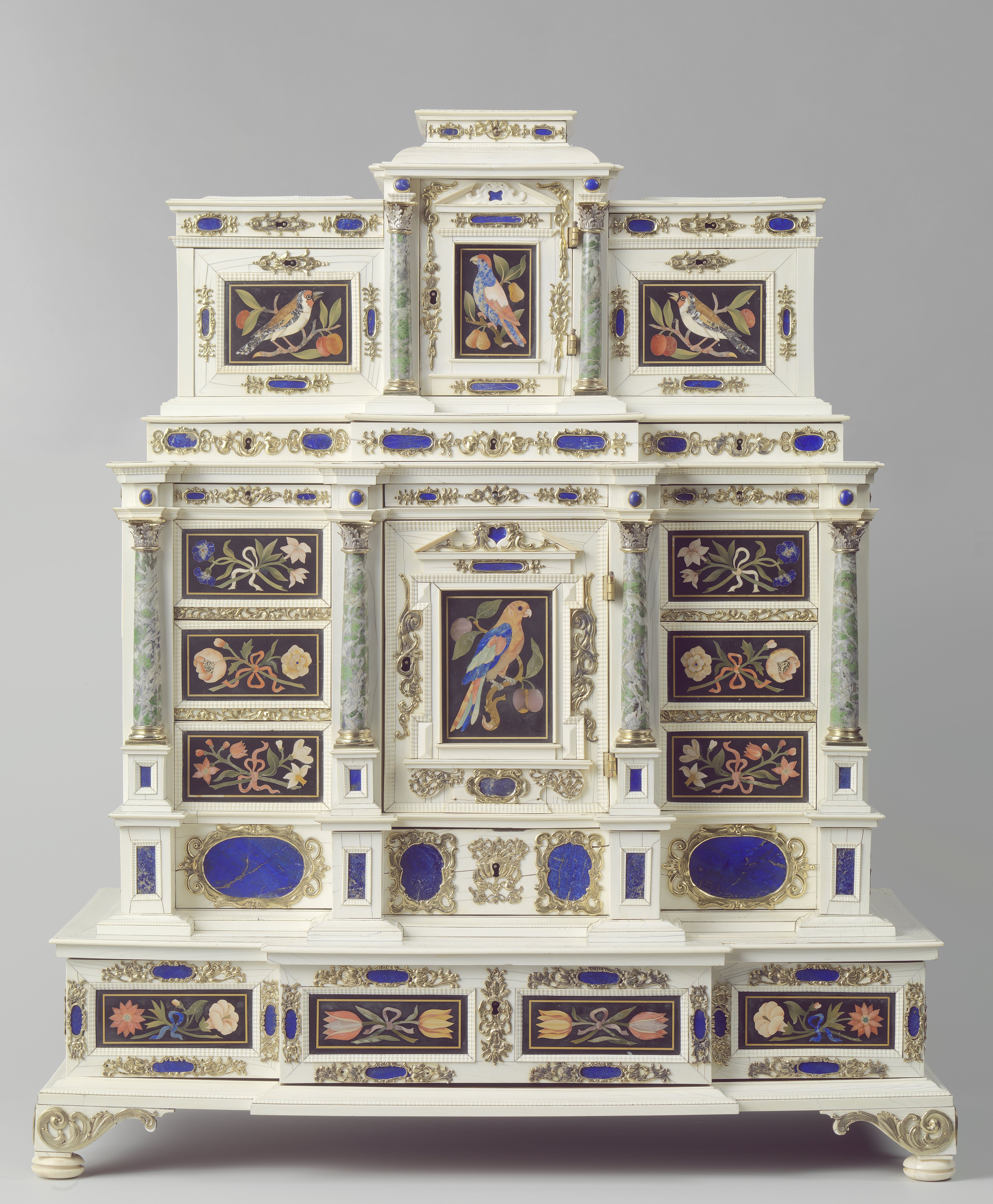}
\label{fig:example_data}
\caption{\textit{Cabinet}, anonymous, c. 1660 - c. 1670. An example of the data in the Rijksmuseum dataset. Among the materials used are oak, maple, lapis lazuli, silver, and copper. The description for the image is "architecture birds branch bunch column flowers fruits ornament pear pillar stick".}
\end{figure}

Zero-shot learning (ZSL) techniques address a type of classification problem where not all classes which may need to be considered at test time have corresponding examples in training \cite{Goodfellow-et-al-2016}. As such, traditional machine learning techniques that rely on generalising from a large quantity of training examples are unable to perform well on these tasks. ZSL instead learns the relationship between provided class \textit{descriptions}, and the content of the training samples. This way, images belonging to an as-yet unseen class can still be classified successfully, so long as a description of what that class might look like exists.

When a museum or gallery obtains a new piece for its collection, it is important to identify the materials with which it was made. This is not only a matter of human interest, but often critical to its preservation, as many materials require specific storage conditions or treatments \cite{Pretzel2003}. With this in mind, we wished to consider a machine learning approach to this problem - would it be possible to train a model to recognise the materials with which a piece was produced from an image of that piece?

While sufficient for common materials such as canvas or bronze, traditional machine learning approaches are unable to handle uncommon materials in this task. As an example, a subset of the dataset comprising only 12,000 images contained 36 examples of materials which were only seen a single time, such as 'fruitwood' or 'nautilus shell'.

Our project employs the Rijksmuseum dataset \cite{Rijksmuseum}. This dataset contains images and descriptions of a large portion of the Rijksmuseum collection, totalling some 350,000 images. Each object in the Rijksmuseum dataset includes a short description of the piece itself. 

Typical ZSL approaches use for the class descriptions English words which directly correspond with features of the image (e.g. \cite{Frome}). Our dataset instead uses English words which \textit{correlate} with the materials used in the piece, but do not directly correspond. This is because the English words contained in the dataset refer to the \textit{subject} of the artwork - a landscape, a self-portrait, etcetera. Our goal is to determine whether this correlated information is sufficient to describe new classes, or whether it is insufficiently direct to achieve good performance.




\section{Methodology}
\label{sec:methodology}
\subsection*{Dataset Overview}
We seek to investigate the Rijksmuseum dataset which contains pictures of all the artworks in the museum, accompanied by a significant amount of metadata. In this project, we rely upon provided English descriptions of the subject of the artworks to identify the artistic medium. A sample of the available data is presented in Figure \ref{fig:example_data}. 

To process the descriptions of the images, we use the Global Vectors for Word Representation (GloVE) algorithm. This transforms the textual descriptions into word embeddings that can be input to the concept embedding layer of our model. 

For each material, we take the set of all words that are used to describe artworks which use that material. From this set, the 25 most frequent words are taken. This list is then used to describe the class in question. 

As recommended by \cite{Frome}, we use a pre-trained network for image feature recognition. In our initial experiments we used VGG16; our first experiments presented in this report also use VGG16, but we in addition show how using ResNet and Xception in the pre-training stage affects our results. 

\subsection*{Fixing Dataset Imbalance}
One of the main issues identified in our dataset was the significant class imbalance. 87\% of our dataset (38642 images) was comprised of images belonging to the class \textit{paper}, while the next most frequent material only comprised 6\% of the dataset. As a result, our initial model was highly biased towards the more frequent materials. Such a significant imbalance can cause serious inaccuracies \cite{Goodfellow-et-al-2016}. To solve this issue, we normalised our dataset to keep a balance between our classes even if this meant a reduction in the number of training images for each class.

As we have built a multi-class model with a dataset comprising of 147 classes, we elected to reduce the size of our dataset and reduce the class imbalance by selecting at most 147 images from each class. In the case where there the original dataset did not have sufficient images available, we selected all images available from that class. The selection was done randomly. After this step, we have considerably improved the structure of our dataset. The most frequent materials are now only comprising 1\% of our dataset each, while the least frequent materials comprise 0.001\% of our dataset. Compared to the previous split, this is a considerable improvement. We investigate later how this change affects the results of our model.

\begin{figure}
\includegraphics[width=0.5\textwidth]{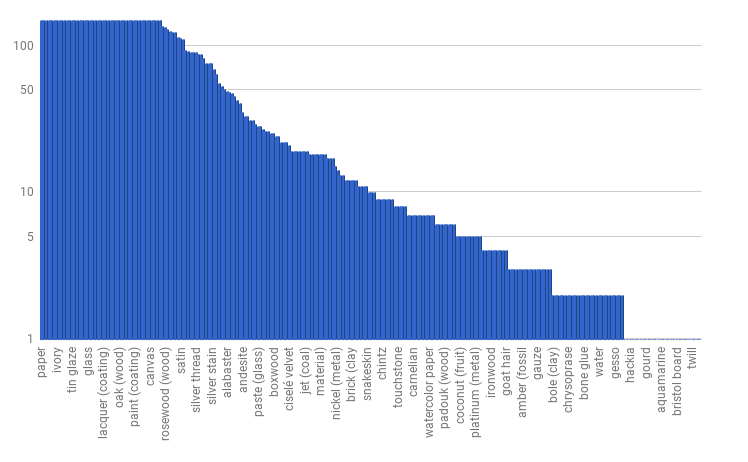}
\caption{Plot of the frequency of occurrence of each material in the dataset (log scale). We can see that our reduced dataset is more balanced while still sparse enough to allow us to explore a ZSL approach.}
\end{figure}

\subsection*{Tate Dataset}
In order to evaluate the quality of the model in recognising unseen classes, we opted to draw our final test data from a distinct dataset. The Tate gallery created a similar dataset to the Rijksmuseum, with 70,000 artworks from their own collection. Even though the artworks are from a different dataset, the format of the data is similar: each artwork has an English description of its subject matter, which does not describe the materials. Additionally there is a list of materials used to create each object, which we use to evaluate the quality of our test-time predictions.

We were able to obtain 5,000 images with corresponding class meta-data from the Tate dataset, which we have held aside as test data for our final model. 

\begin{figure}[h]
\includegraphics[width=0.5\textwidth]{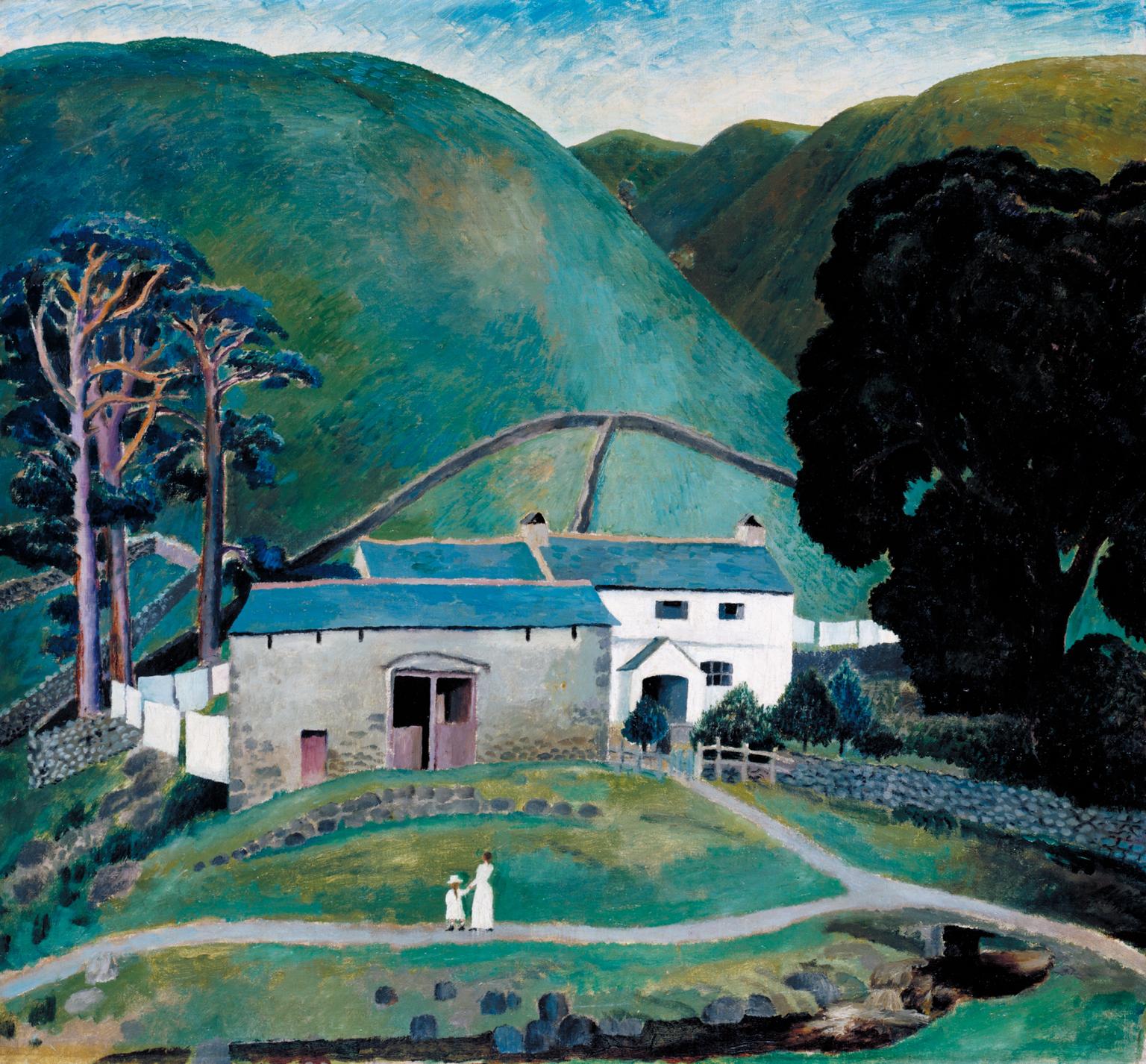}
\label{fig:tate_data}
\caption{\textit{Farm at Watendlath}, Dora Carrington, 1921. An example from the Tate dataset, which contains similar metadata to the Rijksmuseum. The materials for this piece are oil and canvas. The processed description is "adult agricultural architecture barn beck city country county cumbrium district domestic england farm farmland feature field figure hill inland keswick lake landscape laundry line man-made natural nature object path person pine place region stream subject sycamore town townscape tree uk village wall washing watendlath". As with the Rijksmuseum dataset, these descriptions refer to the subject of the piece, and not the object itself.}
\end{figure}

\subsection*{Model Overview}

\begin{figure}
\includegraphics[width=0.5\textwidth]{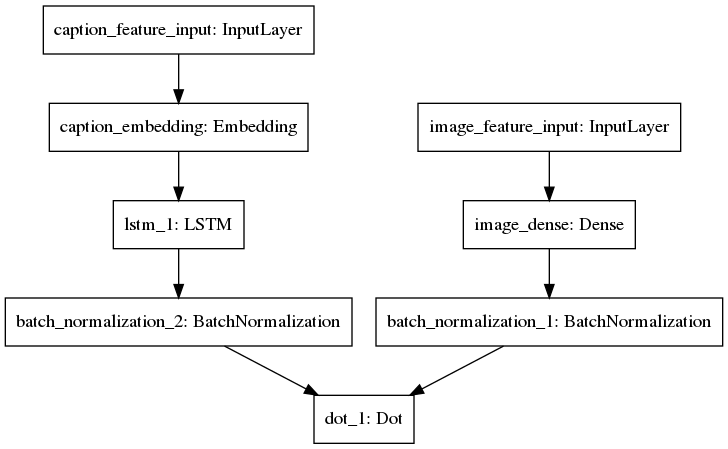}
\label{fig:model}
\caption{Structure of the baseline model, inspired from the DeViSE model published by \cite{Frome}.}
\end{figure}
The Rijksmuseum artwork descriptions are provided as an English paragraph outlining the piece. To process the captions, we split these descriptions into word tokens. Once this has been done, we remove any stop words, such as "the", "a', or "an",  since they do not provide meaningful contextual information. Finally, we encode these tokens as 100-dimensional feature vectors using the GloVE pre-trained word embeddings. 

The next step is to recognise image features. The pre-trained networks encode each image into a 100-dimensional feature vector, which is passed into our model along with the caption embeddings. By using a pre-trained image network to encode our images, we significantly reduce the training time, as we do not need to train a separate model to recognise image features. 

The model presented in Figure 4 represents our baseline model, based upon the DeViSE model published by \citet{Frome}. The image features obtained after pre-processing go through a dense layer followed by a batch normalization layer, while the word embeddings are input to an LSTM layer, again followed by batch normalization. Unlike the original paper, where a concatenation operation is used to combine the two inputs, we combine the two inputs through a normalised dot product. This is done in order to allow the model to learn any viable relationship between image features and description features. When training with a concatenated feature vector layer, the model must instead learn to accurately reproduce this precise pair of vectors, which is a substantially harder task, and therefore liable to produce inferior results at test-time.

\section{Experiments}
We wanted to explore multiple configurations of our baseline model. Additionally, we expressed interest in expanding the depth of our model to see if this improves results. Since our baseline uses a different image model than the one recommended in the original paper describing the DeViSE model, we also aim to investigate the outcome of using different models for image feature extraction. In the end, we will test the model that performs best in all the experiments against the Tate Dataset, described previously. For each of those experiments, we will explain our motivation as well as give detailed information on how the experiment was run and how it can be reproduced. We will present our results and discuss how well they support our hypotheses. Where we give a graphical representation of our results, we apply an exponential smoothing with centre of mass (CoM) = 5, to make the figures more readable. All the results presented in tables or in text show the original data.

\subsection{Hyper-parameters}
We experiment with a wide range of hyper-parameters to finely tune our model and improve performance. All experiments in this section have been run on a validation dataset obtained by randomly selecting 30\% of the total number of classes and excluding them from the training dataset, for 1000 epochs. 

\subsubsection{Optimisers}
\paragraph{Motivation}
We first investigate which optimiser provides the best results for our dataset. Previous research by \citet{pennington2014glove} suggests that AdaGrad is suitable for sparse datasets, as it adjusts the learning rate based on the frequency of the parameter. It has been shown that AdaGrad can be successfully used when training the GloVE model \cite{pennington2014glove}, we expect that AdaGrad will perform better than RMSProp and Adam, which are now more widely used. 
\paragraph{Description}
We train our model using three gradient descent optimization algorithms: AdaGrad, RMSProp, and Adam.

AdaGrad \cite{Duchi2011} works by modifying the learning rate per-parameter. Parameters which are sparse are given much higher learning rates than parameters which are less sparse. For each parameter $\theta_i$ at timestep $t$, the general learning rate $\eta$ is updated as follows:

\begin{equation}
\theta_{t+1,i} = \theta_{t,i} - \frac{\eta}{\sqrt{G_{t,ii}+\epsilon}} \cdot g_{t,i}
\end{equation}

Where $G_{j,j}$ is the diagonal of the outer product matrix: 

\begin{equation}
G = \sum^{t}_{i=1}g_i g_i^\intercal
\end{equation}

Here, $g_i$ is the gradient at $t$. The diagonal is then given by:

\begin{equation}
G_{j,j} = \sum^{t}_{i=1}g^2_{i,j}
\end{equation}

RMSProp \cite{Hinton} divides the learning rate for each weight by an average of the magnitudes of the gradients for that weight, weighted by recency. The first average magnitude is calculated as a mean square:

\begin{equation}
    MS(w,t) = 0.9 * MS(w,t-1) + 0.1 * \bigg( \frac{\partial E}{\partial w}(t) \bigg) ^2
\end{equation}

This results in a moving average of the gradient for the weight. The parameters are then updated:

\begin{equation}
    w = w - \frac{\nu * \nabla w}{\sqrt{MS}}
\end{equation}

Where $\nu$ is the learning rate.

Adam \cite{Kingma2014b} is similar to RMSProp, but averages of the gradients as well as the second moments are maintained. 

The first and second moments are updated much the same as RMSProp:

\begin{equation}
    m_w^{t+1} = \beta_1 m_w^t + (1 - \beta_1) \nabla_w
\end{equation}

\begin{equation}
    v_w^{t+1} = \beta_2 v_w^t + (1 - \beta_2) \nabla_w^2
\end{equation}

Where $m$ is the first moment and $v$ the second moment of the gradient, and $\beta_1$ and $\beta_2$ are the first and second forgetting parameters respectively.

In order to update the weights, adjusted moments are calculated:

\begin{equation}
    \hat{m}_w = \frac{m_w^{t+1}}{1 - \beta_1^t}
\end{equation}

\begin{equation}
    \hat{v}_w = \frac{v_w^{t+1}}{1 - \beta_2^t}
\end{equation}

These adjusted moments are then used to update the weights:

\begin{equation}
    w^{t+1} = w^t - \nu \frac{\hat{m}_w}{\sqrt{\hat{v}_w} + \epsilon}
\end{equation}

Where $\nu$ is the learning rate and $\epsilon$ is a parameter to prevent division by zero.

\paragraph{Results}
The model was trained using a learning rate of 0.001 for 1000 epochs with each optimiser.

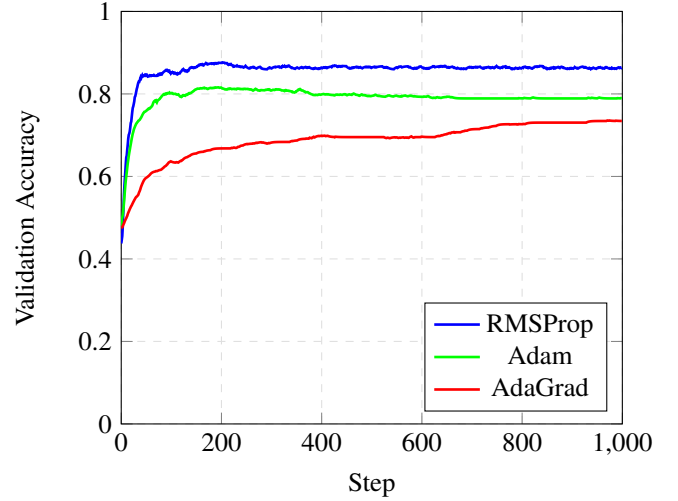
\begin{figure}
  \begin{tikzpicture}
  \begin{axis}[
      xlabel=Step,
      ylabel=Validation Accuracy,
      xmin= 0,
      xmax = 1000,
      ymin = 0,
      ymax = 1,
      grid=major,
      legend pos = south east,
      major grid style={dashed, gray!25},
      width=\linewidth,]
  \addplot[line width=1pt,solid,color=blue,domain=100:1] %
      table[header=true, x=Step,y=Value,col sep=comma]{run_rmsproplr001_tag_val_acc.csv};
  \addlegendentry{RMSProp};
  \addplot[line width=1pt,solid,color=green, domain =100:1] %
      table[header=true, x=Step,y=Value,col sep=comma]{run_adam_tag_val_acc.csv};
  \addlegendentry{Adam};
  \addplot[line width=1pt,solid,color=red, domain = 100:1] %
      table[header=true, x=Step,y=Value,col sep=comma]{run_adagrad_tag_val_acc.csv};
  \addlegendentry{AdaGrad};
  \end{axis}
  \end{tikzpicture}
  \caption{Validation accuracy for different optimisers. Although we hypothesized that AdaGrad would outperform the other optimisers, it is in fact the worst performing of those we tested.}
  \label{fig:optimisers} 
\end{figure}

As can be seen in Figure \ref{fig:optimisers}, AdaGrad performs notably worse than RMSProp and Adam. AdaGrad returned a final accuracy value of $0.734375$, compared to RMSProp which returned $0.8671875$. 


\paragraph{Discussion}
Our results disprove our hypothesis that AdaGrad would perform better on our task than Adam or RMSProp. It is possible that because our model operates on pre-trained image features, the parameters on which the model must train are not sparse. This could explain why the benefits AdaGrad provides on sparse parameters are not seen here. As RMSProp performs significantly above the other optimisers, we use it as our optimiser for the remaining experiments.

\subsubsection{Learning Rates}
\paragraph{Motivation}
Choosing the right learning rate is essential to ensure our model converges. It has been shown that learning rates which are too low can lead to linear improvements \cite{learning_rate}, preventing the model from effectively learning from the data. At the same time, very high learning rates will cause the loss to increase exponentially, while high learning rates will at the start encourage a quick decrease of the loss function, but will eventually prevent the model from converging \cite{learning_rate}. 

\paragraph{Description}
We use a standard range of learning rates at powers of $10$, from $-1$ to $-4$.

\begin{figure}
  \begin{tikzpicture}
  \begin{axis}[
      xlabel=Step,
      ylabel=Validation Accuracy,
      xmin= 0,
      xmax = 1000,
      ymin = 0,
      ymax = 1,
      grid=major,
      legend pos = south east,
      major grid style={dashed, gray!25},
      width=\linewidth,]
  \addplot[line width=1pt,solid,color=blue,domain=100:1] %
      table[header=true, x=Step,y=Value,col sep=comma]{run_rmsproplr1_tag_val_acc.csv};
  \addlegendentry{Learning rate 0.1};
  \addplot[line width=1pt,solid,color=green, domain =100:1] %
      table[header=true, x=Step,y=Value,col sep=comma]{run_rmsproplr01_tag_val_acc.csv};
  \addlegendentry{Learning rate 0.01};
  \addplot[line width=1pt,solid,color=red, domain = 100:1] %
      table[header=true, x=Step,y=Value,col sep=comma]{run_rmsproplr001_tag_val_acc.csv};
  \addlegendentry{Learning rate 0.001};
  \addplot[line width=1pt,solid,color=pink, domain =100:1] %
      table[header=true, x=Step,y=Value,col sep=comma]{run_rmsproplr0001_tag_val_acc.csv};
  \addlegendentry{Learning rate 0.0001};
  \end{axis}
  \end{tikzpicture}
  \caption{Validation accuracy for different learning rates. With the exception of the smallest learning rate, performance is nearly identical across those tested.}
  \label{fig:learning_rates}
  
\end{figure}
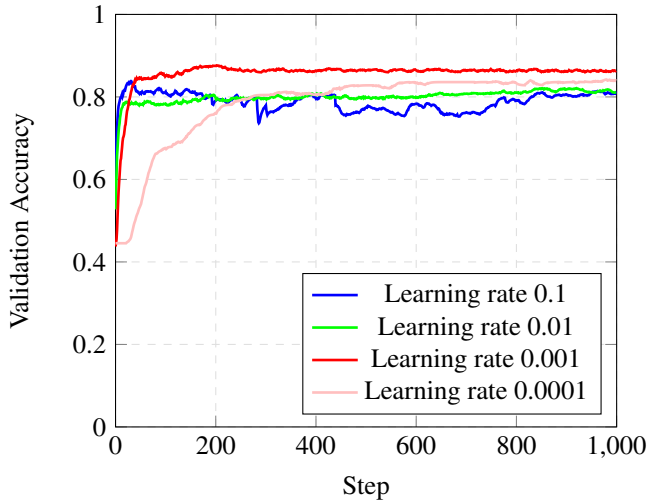

\paragraph{Results}
Figure \ref{fig:learning_rates} shows the results obtained using different learning rates, after training the model for 1000 epochs using the RMSProp optimiser. Although all learning rates achieve reasonably similar performance, a learning rate of $10^{-3}$ achieved accuracy above the other rates tested.


\paragraph{Discussion}
As expected, lower learning rates result in a slower convergence process. While the end results are not significantly different, we can observe that a learning rate of \textit{0.001} results in the best accuracy obtained after the shortest number of epochs. By using this value in further experiments, we can ensure that the model will converge.

\subsubsection{Batch Sizes}
\paragraph{Motivation}
Particularly for larger datasets, selecting an optimal batch size to balance training time with performance is critical to training the network. In theory, the size of the batch is inversely correlated to the performance of the model.
\paragraph{Description}
We used a batch size of 128. This became the basis for our experiments in this section. As a result, we chose to test powers of 2, between 5 and 8.
\paragraph{Results}
As is to be expected, the best performance came from the lowest batch size, 32. However, we recorded high performance from a batch of size 128, and using a larger batch size substantially reduces the training time of the model. As such, we have selected 128 as our batch size for the remaining experiments.
\paragraph{Discussion}
While not returning the best possible performance, we selected a batch size which ensures a reasonable compromise between accuracy and training time. A larger batch means that the weights do not have to be updated as often, speeding up training, but this reduces the performance of gradient descent.
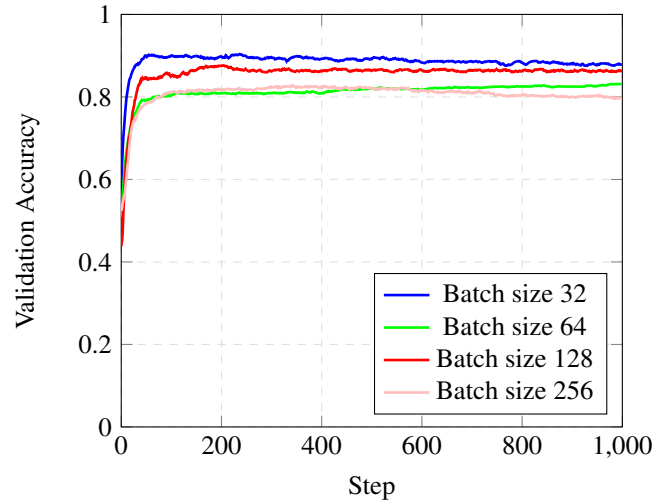
\begin{figure}
  \begin{tikzpicture}
  \begin{axis}[
      xlabel=Step,
      ylabel=Validation Accuracy,
      xmin= 0,
      xmax = 1000,
      ymin = 0,
      ymax = 1,
      grid=major,
      legend pos = south east,
      major grid style={dashed, gray!25},
      width=\linewidth,]
  \addplot[line width=1pt,solid,color=blue,domain=100:1] %
      table[header=true, x=Step,y=Value,col sep=comma]{run_bs32_tag_val_acc.csv};
  \addlegendentry{Batch size 32};
  \addplot[line width=1pt,solid,color=green, domain =100:1] %
      table[header=true, x=Step,y=Value,col sep=comma]{run_bs64_tag_val_acc.csv};
  \addlegendentry{Batch size 64};
  \addplot[line width=1pt,solid,color=red, domain = 100:1] %
      table[header=true, x=Step,y=Value,col sep=comma]{run_rmsproplr001_tag_val_acc.csv};
  \addlegendentry{Batch size 128};
  \addplot[line width=1pt,solid,color=pink, domain =100:1] %
      table[header=true, x=Step,y=Value,col sep=comma]{run_bs256_tag_val_acc.csv};
  \addlegendentry{Batch size 256};
  \end{axis}
  \end{tikzpicture}
  \caption{Validation accuracy for different batch sizes. As expected, smaller batch sizes typically produce higher accuracy.}
  \label{fig:batch_sizes}
  
\end{figure}

\subsection{Model Architecture}
\label{sec:modelarchitecture}
\paragraph*{Motivation}
In order to provide the best possible network performance all hyper-parameters must also be investigated at different network depths. Changing the depth of a network can provide vastly different results as providing more layers provides greater opportunity for the network to learn. 

\paragraph*{Description}
We assess the performance of two different network depths in addition to the base network. The new networks add different numbers of activation layers using \textbf{S}caled \textbf{E}xponential \textbf{L}inear \textbf{U}nit (SELU) \cite{selu}, in between dense layers, as an activation function. All the hyper-parameter experiments described above have been replicated for the new model architectures. As we have already given a motivation and description of how these experiments were done, we will only present the final results for these experiments. When reporting the results, we will list the hyper-parameters used for each experiment in this order: optimizer used, learning rate (LR) and batch size (BS). 

\subsubsection*{Medium Depth}
The medium depth network implemented further layers in addition to the base network layers. The size of the medium network was chosen to balance potential performance improvements with the significant increase in training time due to the additional parameters.

\paragraph*{Results}
The medium depth network architecture performed similarly to the baseline network. However the results (Table \ref{table:meddepth}) prove that the network generally performs worse on the hyper-parameter tests.

\begin{table}
\begin{center}
 \begin{tabular}{||c c||} 
 \hline
 Parameters & Accuracy \\
 \hline\hline
RMSProp, LR = 0.001, BS = 128 & 67.1\%\\ 
 \hline
RMSProp, LR = 0.001, BS = 64 & 77.0\%\\
 \hline
RMSProp, LR = 0.001, BS = 32 & 75.6\%\\
 \hline 
RMSProp, LR = 0.01, BS = 128 & 71.8\%\\
 \hline 
RMSProp, LR = 0.1, BS = 128 & 55.4\%\\
 \hline 
Adam, LR = 0.001, BS = 128 & 77.3\%\\
 \hline 
AdaGrad, LR = 0.001, BS = 128 & 62.5\%\\
 \hline 
\end{tabular}
\end{center}
\caption{Maximum validation accuracy obtained for our medium depth model, with different hyper-parameters.}
\label{table:meddepth}
\end{table}

\subsubsection*{Large Depth}
The large depth network adds layers onto the medium depth network. The network was not made deeper as otherwise the training of the network becomes less effective without more data.

\paragraph*{Results}
While still not performing as well as the baseline network, the network with the largest depth has comparable performance with the medium depth network. The results shown in Table \ref{table:large depth} indicate that neither depth can be labelled as superior to the other as both networks outperform the other with various different configurations.

\begin{table}[htpb]
\begin{center}
 \begin{tabular}{||c c||} 
 \hline
 Parameters & Accuracy \\
 \hline\hline
RMSProp, LR = 0.001, BS = 128 & 73.4\%\\ 
 \hline
RMSProp, LR = 0.001, BS = 64 & 69.1\%\\
 \hline
RMSProp, LR = 0.001, BS = 32 & 78.1\%\\
 \hline 
RMSProp, LR = 0.01, BS = 128 & 70.3\%\\
 \hline 
RMSProp, LR = 0.1, BS = 128 & 63.2\%\\
 \hline 
Adam, LR = 0.001, BS = 128 & 75.7\%\\
 \hline 
AdaGrad, LR = 0.001, BS = 128 & 58.5\%\\
 \hline 
\end{tabular}
\end{center}
\caption{Maximum validation accuracy obtained for our deepest model, with different hyper-parameters.}
\label{table:large depth}
\end{table}

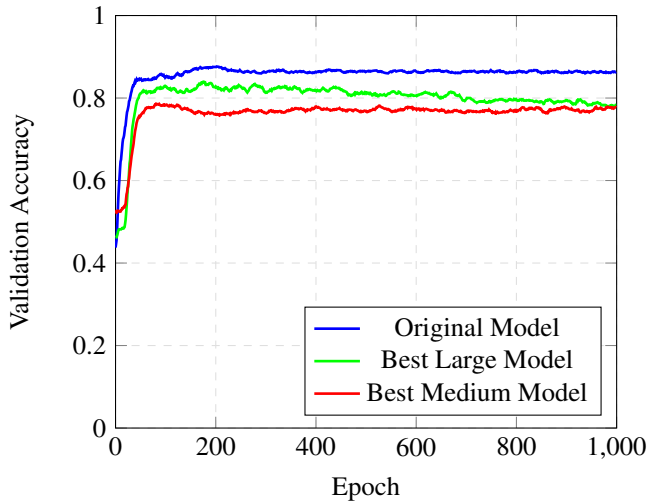
\begin{figure}
  \begin{tikzpicture}
  \begin{axis}[
      xlabel=Epoch,
      ylabel=Validation Accuracy,
      xmin= 0,
      xmax = 1000,
      ymin = 0,
      ymax = 1,
      grid=major,
      legend pos = south east,
      major grid style={dashed, gray!25},
      width=\linewidth,]
  \addplot[line width=1pt,solid,color=blue, domain = 100:1] %
      table[header=true, x=Step,y=Value,col sep=comma]{run_rmsproplr001_tag_val_acc.csv};
  \addlegendentry{Original Model};
  \addplot[line width=1pt,solid,color=green, domain =100:1] %
      table[header=true, x=Step,y=Value,col sep=comma]{run_large_lg_lr_001_opt_rms_bs_32_VGG16_tag_val_acc.csv};
  \addlegendentry{Best Large Model};
  \addplot[line width=1pt,solid,color=red, domain = 100:1] %
      table[header=true, x=Step,y=Value,col sep=comma]{run_medium_med_lr_001_opt_rms_bs_64_VGG16_tag_val_acc.csv};
  \addlegendentry{Best Medium Model};
  \end{axis}
  \end{tikzpicture}
  \caption{Validation accuracy for different model depths.}
  \label{fig:deeper_networks}
  
\end{figure}

\paragraph{Discussion}
Tables \ref{table:meddepth} and \ref{table:large depth} show the results obtained when running experiments with the two new models. In Figure \ref{fig:deeper_networks}, we present a graphical comparison between our best performing configurations for the baseline and the two new deeper models. We can see that our baseline configuration produces significantly better results than the two deeper models. This is likely to do with the specific challenge of our model and dataset. 

\subsection{Pre-training Models}
\paragraph*{Motivation}
As explained in Section \ref{sec:methodology}, instead of building a separate model to extract image features from our artwork dataset, we use a pre-trained network. In the experiments presented so far, we used VGG16 as recommended by \citet{Tejaswin2017}, on whose implementation we base our model. While the initial results of our model are satisfactory, we explore how different pre-training models affect the classification in order to investigate whether an alternative model may produce better results. Much of the previous research done on artistic datasets (e.g. \citet{2017recArtStyle}) is performed using ResNet50 for image feature extraction. This suggests that the ResNet50 model might be better suited for our dataset. 

In addition, the original DeViSE model uses Xception in the feature extraction step \cite{Frome}. However, their model is not applied to an artistic dataset, where the image content can be highly abstract. Since Xception has been successfully used previously in image classification with very small dataset problems \cite{DBLP:journals/corr/Chollet16a}, it is possible that this approach will give better results.

\paragraph{Description}
In this experiment, we want to understand which pre-training models yield the best results. We run the experiments on the best performing original, medium and large depth model (Section \ref{sec:modelarchitecture}). We use VGG16, ResNet50 and Xception as pre-training methods and train the ZSL classifier for 1000 epochs. The validation is performed on the same dataset for each method. 

\paragraph{Results}
Table \ref{table:pretraining} shows the results obtained for our experiments. The best result is obtained using our best baseline configuration, using VGG16, with an accuracy of 86.3\%. However, as the model gets more complex, VGG performs worse, while Xception consistently gives good results. 
\begin{table}
\begin{center}
 \begin{tabular}{||c c c c||} 
 \hline
 Model Architecture & VGG16 & ResNet50  & Xception\\
 \hline\hline
 baseline & 86.3\% & 70.8\% & 85.1\% \\ 
 \hline
 large & 78.13\% & 69.0\%  & 82.9\%\\
 \hline
 medium & 77.0\% & 69.2\% & 80.2\% \\
 \hline 
\end{tabular}
\end{center}
\caption{Maximum accuracy obtained for different model architectures, using VGG16, ResNet50 and Xception for image feature extraction.}
\label{table:pretraining}
\end{table}

\paragraph{Discussion}
While research shows that ResNet50 is a good model when working with artistic datasets, our experimental results contradict this hypothesis. We believe this might be due to the sparsity of the dataset. In fact, the Xception model, which has been shown to perform well on sparse datasets, provides consistently good results, regardless of the complexity of the model. The initial implementation using VGG16 still gives the highest validation accuracy value, but this is not significantly higher than the value obtained using Xception on the same model. 


\section{Testing}
To asses the final performance of the network, a new as-yet-unseen dataset had to be used as a testing set. We use the Tate Museum dataset to achieve this task by training a network on the Rijksmuseum dataset and then using it to predict material labels for the Tate Museum data. From the hyper parameter experiments we select the network with the best performance in the experiments to test the Tate dataset on.

\subsection*{Testing Results}
When evaluating 5,000 images from the Tate dataset on our best performing model, we obtained a classification accuracy of 48.42\%. Given that the Tate dataset sample used contains 363 distinct materials across all the images (more than double the 147 materials in the Rijksmuseum dataset), we believe that this result indicates that the method is viable for reasoning about an entirely unseen collection.

\section{Related Work}
This project combines several different techniques to create a unique and novel approach to an interesting problem. As a result of combining multiple techniques there is a large quantity of work that relates to this project. 

\subsection*{Recognizing Art Style}
The first paper discussed here investigates deep learning approaches to classifying artwork styles using convolutional neural networks \cite{2017recArtStyle}. The paper explores classification using pre-trained networks. Specifically it investigates the performance of AlexNet \cite{AlexNet} and ResNet50 in the artistic style classification task. To test the accuracy of their model, the Wikipainting dataset was used, which is a large dataset collected from WikiArt. This approach reported a classification accuracy of 62\% which increased to 93\% when using Top-5 accuracy.

Another paper learned representations of artistic styles to transfer images between styles \cite{LRAS}. The paper states that previous processes were very computationally expensive, and propose a convolutional neural network that tackles this problem. In order to test their style transfer network they use ImageNet.

\subsection*{Material Recognition}
There is also great interest in material recognition in the real world. One paper investigated the use of different convolutional neural networks to extract features relating to material classification \cite{idfmr}. This work was motivated by an observed close link between object recognition and material recognition. A new Extended Flickr Material Database (EFMD) was introduced, which was then used to train a convolutional neural network. Using this, a classification accuracy of 82\% is reported, using deep features extracted from multiple pre-trained convolutional neural networks.

Multiple papers address the problem of real world material recognition. In \cite{minc} the authors explain the importance of a well sampled training data set. The Materials In Context Database (MINC) is introduced, which is large and well sampled. Using this dataset to train a convolutional neural network the authors report a material prediction accuracy of 85\% when categorising \textit{patches} of images. The paper also attempts to classify the material contained within each individual pixel of an image. When testing this method the authors managed to build a network that produces 73\% classification accuracy. 

\subsection*{Rijksmuseum Dataset}
The Rijksmuseum dataset also has work directly related to it. \citet{RijkChallenges} presents multiple challenges relating to the classification of data from the Rijksmuseum's dataset. One of these challenges specifically relates to the classification and recognition of materials used in the construction of the artwork. This paper reports a material classification accuracy of 91\% when using 103 labels to classify the entire dataset.

One paper that attempts these challenges and reports state-of-the-art results is the OmniArt paper \cite{OmniArt}. This paper investigates the lack of suitable meta-data held by museum datasets and uses convolutional neural networks to generate meta-data for artworks. Among other features this network generates multi-label material labels for an artwork based on the image. For the same number of labels OmniArt reports 93\% material classification accuracy.
\section{Conclusion}
From the results we conclude that multi-class ZSL is a viable method of predicting material labels for artworks of different mediums. This does require that a dataset is relatively balanced as otherwise the predictions are skewed in favour of the most frequent classes during training. Balancing is achieved by use of a script that limits the relative proportion of examples of each material class. The results obtained from these experiments show lower accuracy values than those reported in \cite{OmniArt} and \cite{RijkChallenges}. However, these models are single-class in contrast to our multi-class approach.

The project investigated several configurations of hyper-parameters in order to determine an optimum setup for the network. As well as this, three different network depths were investigated. From the hyper-parameter experiments, we can deduce that RMSProp with a learning rate of \textit{0.001} and a batch size of 128 will yield the best results in the shortest training time. Our deeper network experiments prove that adding more complexity into the model does not necessarily improve the results. By experimenting with different pre-training methods, we determined that VGG16 provides marginally better results on smaller models, while Xception consistently achieves good validation accuracies values regardless of the model architecture. In contrast, ResNet50 does not perform well on our dataset.

After testing on the Tate data set it was concluded that multi-class ZSL is a viable method of classifying the materials that make up artworks of different mediums using descriptions of the piece. Furthermore, we propose that this approach is valid for multiple different artwork datasets. The results of the models tested on the Tate museum dataset suggest that it is possible to obtain a reasonably accurate classification even when the dataset does not conform to the exact same description format. 

Our research question was \textit{"would it be possible to train a model to recognise the materials with which a piece was produced from an image of that piece?"}. We have shown conclusively that, using a ZSL approach, this is possible - even more so, we have demonstrated that it is possible to apply understanding of one museum's collection to the same task at another museum.

\subsection*{Future Work}
In order to further improve the performance several improvements could be made. Firstly, if a suitable dataset with a defined class hierarchy is used then performance could be compared to \citet{Frome} by using Top-K Hierarchical Accuracy. This would give a better indication of where the project stands in relation to state-of-the-art systems such as \citet{Frome}. 

Other improvements might be to train the original model using more word embeddings by expanding the words allowed in the embeddings. Furthermore, including different meta data from the Rijksmuseum data might provide more meaningful embeddings in the context of other datasets and thus improve the classification accuracy. Failing this, using a dataset with descriptions of a nature more similar to the Rijksmuseum could potentially provide better classification results.

In addition to this, future steps could include training the network with less data to find the minimum quantity of training data which still provides reasonable results. This would potentially extend the use of the project to private artwork stashes and similar collections. This is desirable as without an easy tool to generate data from the artworks gives a greater chance that these artworks will be preserved in digital form.


\bibliography{AOACTBmlp}

\end{document}